# Web-based Teleoperation of a Humanoid Robot

*Chien-Liang Fok, Fei Sun, Matt Mangum, Al Mok, Binghan He, and Luis Sentis*

**Abstract**

*The Cloud-based Advanced Robotics Laboratory (CARL) integrates a whole body controller and web-based teleoperation to enable any device with a web browser to access and control a humanoid robot. By integrating humanoid robots with the cloud, they are accessible from any Internet-connected device. Increased accessibility is important because few people have access to state-of-the-art humanoid robots limiting their rate of development. CARL's implementation is based on modern software libraries, frameworks, and middleware including Node.js, Socket.IO, ZMQ, ROS, Robot Web Tools, and ControlIt!. Feasibility is demonstrated by having inexperienced human operators use a smartphone's web browser to control Dreamer, a torque-controlled humanoid robot based on series elastic actuators, and make it perform a dual-arm manipulation task. The implementation serves as a proof of concept and foundation upon which many advanced humanoid robot technologies can be researched and developed.*

# 1. Introduction

Whereas traditional industrial robots are physically dangerous to be around, humanoid robots are meant to operate in our environments with safety guarantees. The ability to work alongside humans in our homes and workplaces significantly increases the range of applications of humanoid robots. Despite this, their current capabilities are extremely limited [Ackerman 2015]. While the technical challenges are undoubtedly high, another fundamental reason for the relatively low rate of development is the fact that high performance humanoid robots are expensive resulting in few people having access to the hardware. Without access to the hardware, the vast majority of students and learners interested in robotics cannot easily contribute to new practical methods and experimental advancements. In this paper, we address this problem by presenting a software architecture that integrates humanoid robots and web-based teleoperation to provide humanoid robotic learners more access to the hardware.

Humanoid robot teleoperation can be conducted via the Internet and is thus accessible from virtually anywhere. We leverage this fact to enable greater access to state-of-the-art humanoid robot hardware. Our software architecture is called the Cloud-based Advanced Robotics Laboratory (CARL). Using CARL, a user can remotely perform dual arm manipulation via a smartphone's web browser while collecting and visualizing data about the robot's state. This demonstrates feasibility; high performance humanoid robots can be accessed and controlled via a web-based user interface. The focus of this paper is on developing a software architecture that leverages the cloud to increase access to state-of-the-art humanoid robots.

Remotely controlling a humanoid robot over the internet requires overcoming several challenges. First, humanoid robots are highly redundant containing far more Degrees of Freedom (DOFs) than what a human operator will typically want to control. Having redundancy is important however since most humanoid robots must exhibit fast reflexes and juggle numerous objectives like maintaining a particular posture while performing two arm manipulation. This is typically achieved using sophisticated Whole Body Control (WBC) algorithms. Second, WBC algorithms typically require predictable 1kHz servo frequencies. This is difficult to achieve over the Internet due to unpredictable latency and limited bandwidth. For this reason, the WBC process will continue to execute on a computer within the robot. Careful consideration is necessary to limit the amount of data to transfer between the robot, the cloud infrastructure, and the user's device. Finally, the human operator may use a wide variety of devices ranging from relatively powerful desktop computers with large screens to relatively weak mobile devices with small touch-screens. CARL's software architecture is designed to address these challenges.

The remainder of this paper is organized as follows. Section 2 presents related work. Section 3 presents CARL's software architecture. Section 4 presents CARL's user interface. Section 5 presents implementation details. Section 6 presents a usage demonstration. Finally, Section 7 contains conclusions and a discussion of future work.

## 2. Related Work

There are several areas of related work. They include telerobotics, cloud robotics, whole body controllers, and humanoid robot user interfaces. Each are now discussed.

**Telerobotics.** Telerobotics [Goldberg and Siegwart 2001] consists of a human operator controlling a robot over a communication link. The human issues a command that is sent to the robot. The robot executes the command and sends back sensory information, which may include position, video, audio, and haptic data [Willaert 2010, Buys 2011, Bellens 2011, Poorten 2012]. The operator uses this sensory information to issue the next command to the robot. Since the operator has fine-grained control over a robot and the robot typically has very little autonomous decision making capability, the main focus is on operator experience. For this reason, telerobotics sometimes use a dedicated high bandwidth and low latency link that enables the operator to continuously monitor the robot at high fidelity and quickly adapt to changes in the robot's context [Ross 2007, Dudley 2014]. To prevent "move and wait" control style, communication latencies had to remain below 480ms. In addition, video resolutions of 40 pixels per degree is needed for high performance teleoperation.

Teleoperated robotics is in widespread use. Examples include controlling rovers on Mars, fixing the Deepwater Horizon oil well, nuclear material handling, bomb disposal, telesurgery, telepresence, and the recent DARPA Robotics Challenge [Website - DRC]. Recent research projects include fixed-based manipulation [Taylor 1995], mobile manipulation [Osentoski 2012, Pitzer 2012], gardening [Goldberg 2001 Garden], maze navigation [Crick 2011], human-robot interaction research [Toris 2014], and imitation learning [Chung 2014]. The knowledge about user experience parameters gained from telerobotics research may be leveraged to further improve CARL's ability to enable remote access to humanoid robots over the cloud.

**Cloud Robotics.** Cloud robotics [Hu 2012, Kehoe 2015] is a natural extension of telerobotics while its focus is on exploiting the vast amounts of computational resources available on the cloud. Specifically, the goal is to augment the computational resources that are onboard the robot with Internet server farms, data centers, humans, and other robots. Higher levels of computational resources can result in more artificial intelligence and autonomy by enabling large amounts of data and past experiences to be stored, recalled, and interpreted [Pratt 2015].

The paradigm of running part or all of an application on Internet servers is called "cloud computing." There are three levels of cloud computing. The lowest level is Infrastructure as a Service (IaaS), which provides raw computational resources. Examples include Amazon Web Services [Website - AWS], specifically the [Elastic Compute Cloud (EC2)](#) and [Simple Storage Service (S3)](#). The middle level is Platform as a Service (PaaS) where the service includes an operating system and middleware. An example is [Google's Application Engine](#). Finally, the highest level consists of Software as a Service (Saas) where entire applications are provided. Examples include Google's gmail, drive, and calendar. Cloud robotics is a specialization of cloud computing to the robotics field and is still in its infancy. Research projects include DAvinCi (Distributed Agents with Collective Intelligence) [Arumugam 2010] for navigating large environments, RoboEarth for enabling experiences to be shared among successive generations of robots [Waibel 2011, Riazuelo 2015, Website - RoboEarth], and the use of Google's image recognition engine and 3D model data storage service to enable a PR2 robot to recognize and grasp 3D objects in an unstructured environment [Kehoe 2013]. On the commercial front, [Rapyuta Robotics Co., Ltd.](#) is a startup company founded in July 2014 that is focusing on the development of cloud-based multi-robotic systems for security and inspection applications.

The key difference between cloud robotics and CARL is its focus on humanoid robots. Humanoid robots require whole body controllers for fast reflexes and to achieve multiple simultaneous objectives. CARL integrates whole body controllers and web application technologies to enable remote access to state-of-the-art humanoid robots. CARL provides the basic foundation for connecting a user with a smartphone to a humanoid robot. The aforementioned benefit of access to vast computational resources can be added to CARL in the future.

**Whole Body Controllers.** Whole body controllers are particularly useful for humanoid robots due to their relatively high number of joints, which typically range from 16 to over 50, and need for fast reflexes [Website - IEEE RAS Whole-body Control Technical Committee]. They are a type of Multi-Input-Multi-Output (MIMO) controller that takes a holistic view of the robot to achieve multiple simultaneous objectives and constraints like maintaining balance or particular posture while reaching for an object. Servo frequencies on the order of 1kHz enables quick reflexes. Whole body controllers reduce the cognitive load of the operator by providing a higher level of abstraction [Neo 2007]. For example, instead of specifying commands for every joint in the robot, by using a whole body controller the operator can focus on just the 6-DOF Cartesian configurations of the robot's end effectors [Sian 2004]. This reduction in complexity makes remote operation of a humanoid robot using a small device like a smartphone tractable.

Many whole body control algorithms exist and can generally be classified as being optimization-based or analytical. Optimization-based algorithms typically rely on quadratic programming to find a solution to the control problem [Kuindersma 2014, Kuindersma 2015]. Analytical algorithms typically rely on matrix pseudoinverses and null space projections to find a solution [Sentis 2007]. Of these, analytical algorithms are computationally simpler and thus have lower latencies but do not support inequality constraints. CARL's current implementation uses ControlIt! [Fok 2015], a high performance open source implementation of an analytical algorithm called Whole Body Operational Space Control [Sentis 2010, Sentis 2013]. In the future, ControlIt! and thus CARL may be updated to support other whole body control algorithms.

**Humanoid Robot User Interfaces.** Significant amounts of research exists on the user interface for remotely operating a humanoid robot [Nishiyama 2003, Takubo 2004, Vanthienen 2013]. Examples include rFSM [Klotzbucher 2012], iTaSC [Vanthienen 2011], and Robot Task Commander [Hart 2014 RTC]. Affordance templates are often used to specify the location and orientation of physical objects in the environment that the user wants the humanoid robot to manipulate [Hart 2014, Hart 2015]. Exotic input devices like a marionette [Takubo 2004, Takubo 2006], 3D mouse [Nishiyama 2003], goniometers [Castellanos A. 2008], and brain-computer interfaces [Bryan 2011, Finke 2012, Finke 2013] were also studied in the context of enabling remote operation of humanoid robots. These may be integrated into CARL in the future to further improve human-computer interaction.

User interfaces on everyday internet-enabled devices such as smartphones [Brown 2013, Ahn 2014] have been previously used. But these prior works relied on inverse kinematic control for remote operation while CARL implements compliant control through WBC. Another key difference lies in the delivery of the user interface through a web browser. To our knowledge, previous humanoid robot user interfaces were implemented in stand alone applications that required the installation of specialized software. CARL's user interface is provided through a web browser meaning anyone can access the humanoid robot by simply navigating to a particular website. In the future, ideas and lessons derived from the past humanoid robot user interfaces can be integrated into CARL to achieve higher quality user experience.

## 3. CARL Software Architecture

CARL's software architecture is shown in Figure 1. It is designed to enable remote users to access humanoid robots via web browsers. This is achieved by encapsulating each robot within a CARL Operating System (CARLOS), which communicates with a CARL Server located in the cloud. The CARL Server implements a web portal through which users can access the robots via web browsers. Using this architecture, users can access robots from any Internet-connected device since the CARL Server is located in the cloud.

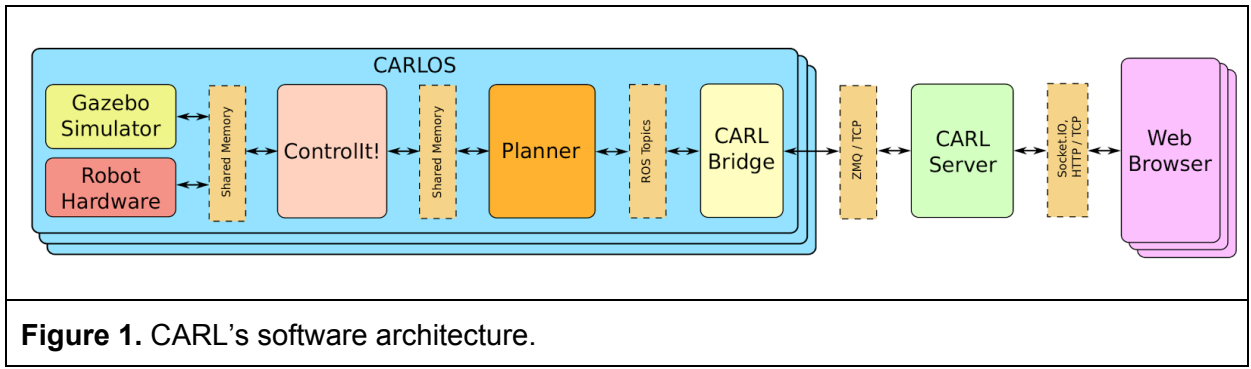

**Figure 1.** CARL's software architecture.

CARLOS is a key component in CARL's architecture that connects the robot to the CARL Server. The robot itself can be real or simulated. Users can rapidly switch between the two for testing and debugging purposes. The robot is connected via shared memory to ControlIt!, which is a framework for controlling many DOF robots and implements the whole body controller. The whole body controller's servo loop consists of the robot sending ControlIt! its latest joint state information, and ControlIt! sending the robot the next joint commands. Shared memory is used to achieve acceptable levels of communication latency and real-time predictability, which are needed to ensure whole body controller stability.

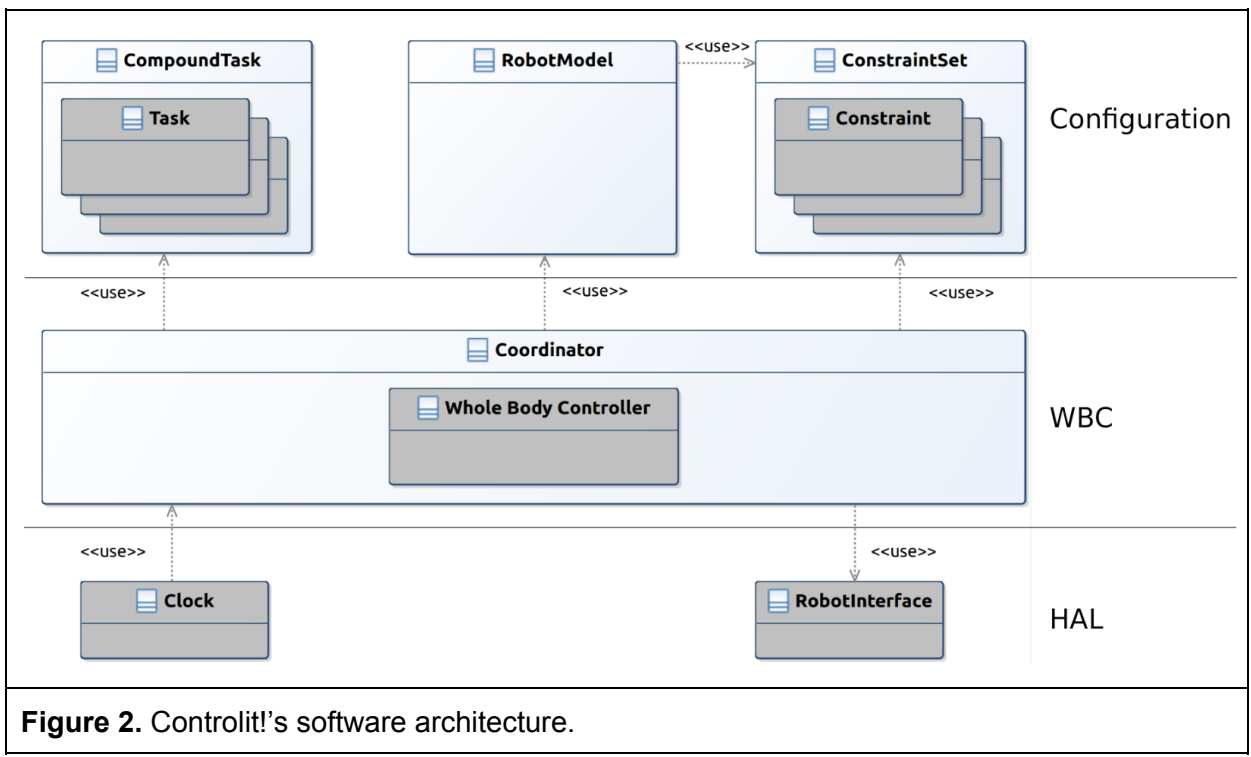

**Figure 2.** Controlit!'s software architecture.

ControlIt!'s software architecture is shown in Figure 2. We highlight a few components that are most relevant to CARL. For full details, refer to [Fok 2015]. ControlIt!'s architecture is divided into three levels: configuration, Whole Body Control (WBC), and Hardware Abstraction Layer (HAL). The configuration level consists of a compound task, a robot model, and a constraint set. The compound task contains a prioritized list of tasks, each specifying a unique objective of the whole body controller. Each task operates in its own space, which typically contains far fewer DOFs than the number of joints in the robot. For example, common tasks include 6-DOF Cartesian position and orientation control of a point on the robot. Offering a lower dimensional space within which to operate is important because it enables remote control of the robot via a small device like a smart phone feasible.

Another notable component within ControlIt!'s architecture is the constraint set. It contains constraints that specify physical limitations of the robot. Examples include joints that are mechanically coupled, or places where the robot contacts the environment. The constraint set can be changed over time to reflect the current configuration of the robot with respect to its environment.

The WBC layer implements the actual whole body controller. The HAL contains a clock and robot interface for enabling different types of robots and real-time operating systems to be supported.

ControlIt!'s compound task and constraint set are connected to a planner via shared memory. The planner provides a sequence of references for the tasks to follow, and can update the constraint set based on changes in the robot's relationship to its environment. The planner ensures the reference trajectories being followed by the whole body controller are smooth. This is necessary since any discontinuities in the trajectory may result in sudden movement of the robot and instability. Shared memory is used between the planner and ControlIt! to ensure low and predictable communication latencies, which is important for real-time operation and controller stability.

The planner is integrated with the Robot Operating System (ROS), a component-based middleware based on communication channels called ROS Topics [Quigley 2009]. ROS is designed to work within a single local area network. To bridge the gap between ROS and the CARL server located in the cloud, CARL implements a CARL Bridge component. The CARL Bridge translates between ROS topics and a globally routable communication channel. Since the bandwidth, latencies, and reliability of the Internet are less predictable, the CARL Bridge includes some intelligence to reduce the amount of data transmitted to the CARL Server.

The CARL Server implements a web portal that is viewable by a wide range of Internet-connected devices. Users directly interact with this portal to gain access to the robot.

# 4. User Interface Design

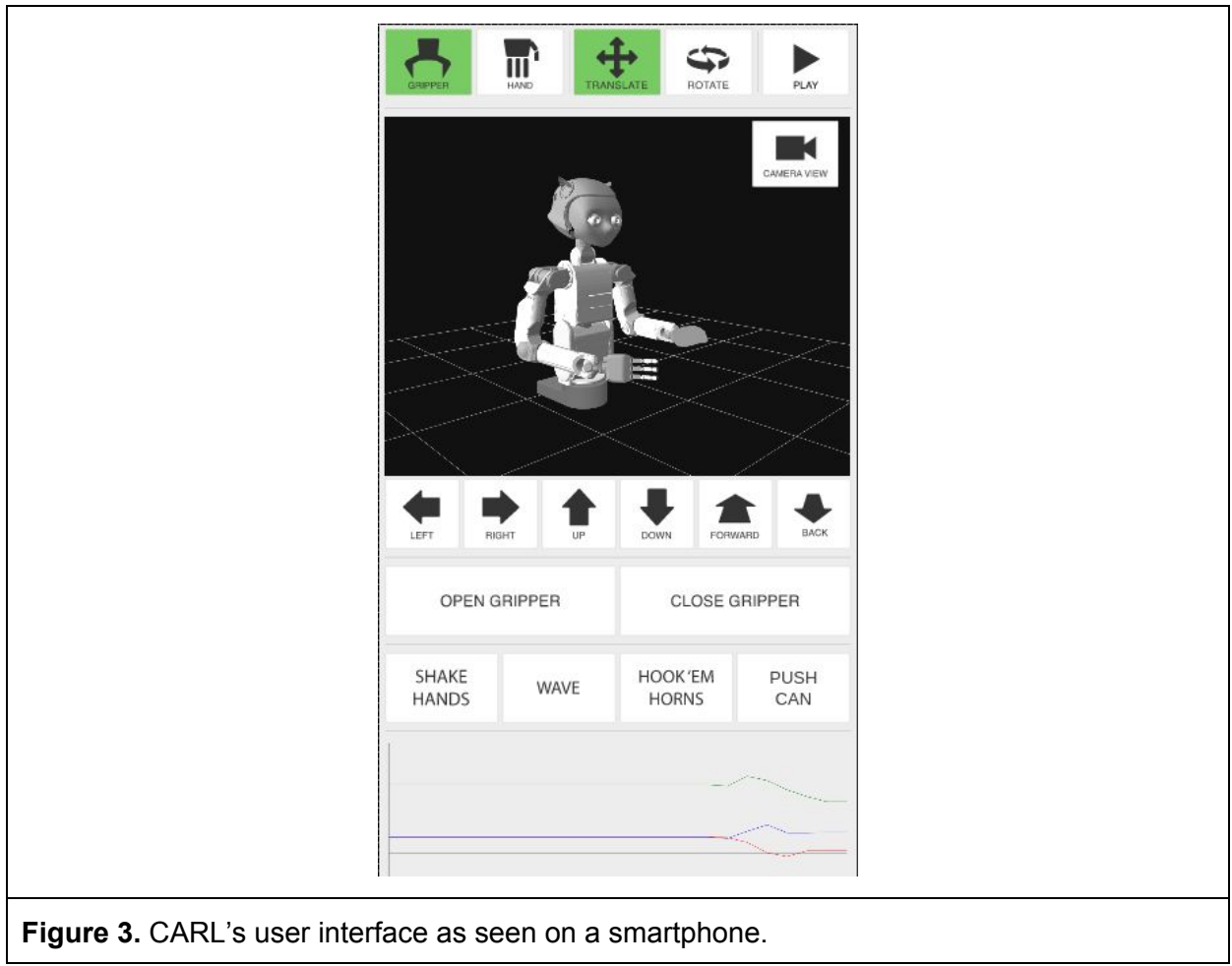

**Figure 3.** CARL's user interface as seen on a smartphone.

A screenshot of CARL's web portal is shown in Figure 3. The user interface is designed to work on the small screen of a smartphone. At the very top are buttons for turning the robot on and off, selecting which end effector to control, and selecting whether to control the selected end effector's position or orientation. Figure 3 shows that the gripper's position is currently under control.

To know the robot's current state, the largest section of CARL's user interface consists of a 3D visualization of the robot. This section can be swapped by webcam feeds to get a video of the robot and its surroundings. The 3D visualization and webcam footage is sufficient for the user to perform telemanipulation tasks. In the future, other more advanced visualizations like affordance templates [Hart 2015], finite state machines [Hart 2014 RTC], and foot step sequences [Stumpf 2014] may be added.

Below the visualization are six buttons that enable a user to change the reference position and orientation of the selected end effector, and open and close the end effector. Each time the user hits one of these buttons, a command is sent by the CARL Server to the planner causing the robot to move a certain increment. Continuous movements are not possible due to high and unpredictable communication latencies. CARL currently provides a "stop-and-wait" style of telemanipulation.

To avoid "stop-and-wait" telemanipulation behavior, CARL's user interface includes a section with buttons that command the robot to perform certain autonomous behaviors. The sophistication of the autonomous behaviors will change over time as technology improves. Currently, autonomous behaviors include things like waving to a person, shaking hands with a person, and executing a University of Texas "Hook'em Horns" gesture, as shown in Figure 4.

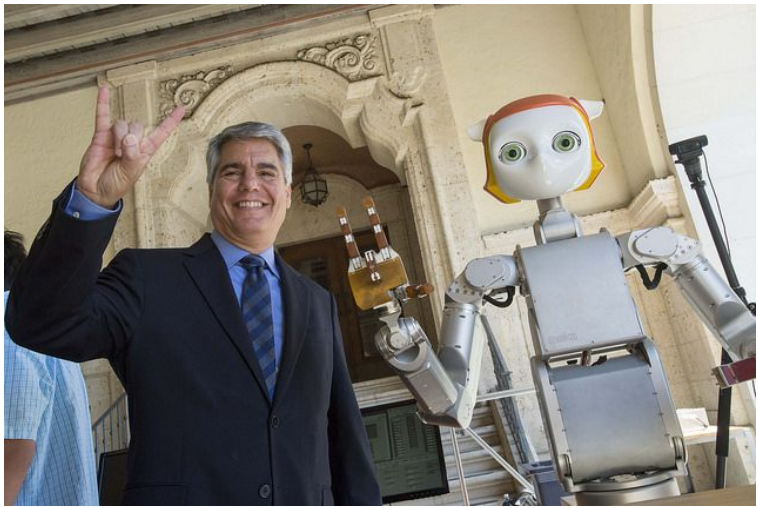

**Figure 4.** CARL being used to make Dreamer, a humanoid robot, autonomously execute a Hook'em Horns gesture with UT President Gregory L. Fenves.

The bottom of CARL's user interface contains a data visualizer. This shows a live stream of data about the robot that is being stored in the cloud. Advanced data analytics can be applied to this data to glean useful information about the robot. For example, it can be used to debug the system [Annable 2013], identify potential health problems with the robot [Pettersson 2004] or ways the controller can be improved [Peters 2003, Ogawa 2014].

The current user interface is tailored specifically to Dreamer, our torque controlled humanoid robot that uses series elastic actuators. In the future, we plan on generalizing our system to support other humanoid robots. In addition, CARL's current user interface only supports one robot at a time. In the future, we plan on improving it to support multiple robots.

# 5. Implementation Details

The CARL user interface is implemented in HTML, CSS, and JavaScript. It uses web standards that are supported by all modern web browsers eliminating the need to install 3rd party plugins like Flash or Java. The web portal is implemented on top of the [Express](#) web application framework, which runs on a [Node.js](#) server.

Communications between the web browser and CARL server are done using [socket.io](#), which is a wrapper around many transport protocols. The selection of which transport protocol to use is done transparently. Websocket is the preferred protocol when it is supported by the browser, which was the case in our system.

Numerous open-source JavaScript libraries are used to achieve a user interface with high quality user experience. Some of the notable ones are:

1. [three.js](#) - a JavaScript 3D library that's used for the 3D visualization of the robot.
2. colladaLoader.js and colladaLoader2.js - JavaScript libraries for loading [COLLADA](#) (COLLAborative Design Activity) files that are part of the robot model.
3. STLLoader.js - A JavaScript library for loading [STL](#) (STereoLithography) files that are part of the robot model.
4. [roslib.js](#) - The core JavaScript library that's part of [Robot Web Tools](#) [Toris 2015] that enables a web browser to interact with ROS. Robot Web Tools provides a collection of open-source modules and tools for building web-based robot applications.
5. [ros3d.js](#) - The core JavaScript library that's part of Robot Web Tools and provides software abstractions for supporting interactive markers, point cloud streaming, and robot visualization.
6. [jsmpeg.js](#) - A MPEG2 decoder written in JavaScript for supporting the webcam within CARL's portal.
7. client.js - The core CARL client JavaScript library.

In the list above, client.js is the main JavaScript library that implements code specific to CARL. It obtains the model of the robot, instantiates a 3D visualization object, a webcam object, and handles all user commands from the browser. It also takes real-time data from the robot like joint state information and visualizes it in the browser.

For the webcam video feed, the CARL user interface creates a WebSocket connection to the webcam server. It then instantiates a jsmpeg video player, which displays the data from the webcam. This goes into the HTML5 canvas component within the client's webpage. By using HTML5, CARL avoids the need to install a specialized video player plugin into the browser.

CARL uses Robot Web Tools for 3D visualization of the robot's model. Robot Web Tools is tightly integrated with the Robot Operating System (ROS). In addition to 3D visualization, it also provides [rosbridge_suite](#) for translating between JSON and ROS messages and thereby connects web browsers to ROS. However, CARL does not use rosbridge_suite because it requires the web server and robot controller to be colocated within the same local area network due to a ROS limitation. CARL overcomes this limitation by implementing carl_bridge, which uses ZMQ to connect between ROS and the web server. This allows CARL to locate its web server anywhere in the cloud instead of in the same local area network as ROS and ControlIt!. Having this capability is important because it enables CARL to scale and support multiple robots each running within their own ROS network.

The planner within CARLOS is written in Python and implements two types of interpolators. For the telemanipulation application, a trapezoidal velocity trajectory generator is used to ensure smooth movements each time the user changes the reference position or orientation of an end effector. For the fully autonomous behaviors, a cubic spline trajectory generator is used to interpolate between the trajectory's waypoints.

# 6. Evaluation

We demonstrate the feasibility of CARL's architecture by allow members of the public to use a smartphone to control Dreamer, as shown in Figure 5. The excellent user experience and intuitive user interface is demonstrated by the fact that untrained users could immediately operate Dreamer using a smartphone and perform basic manipulation tasks even when not within direct line of site with the robot.

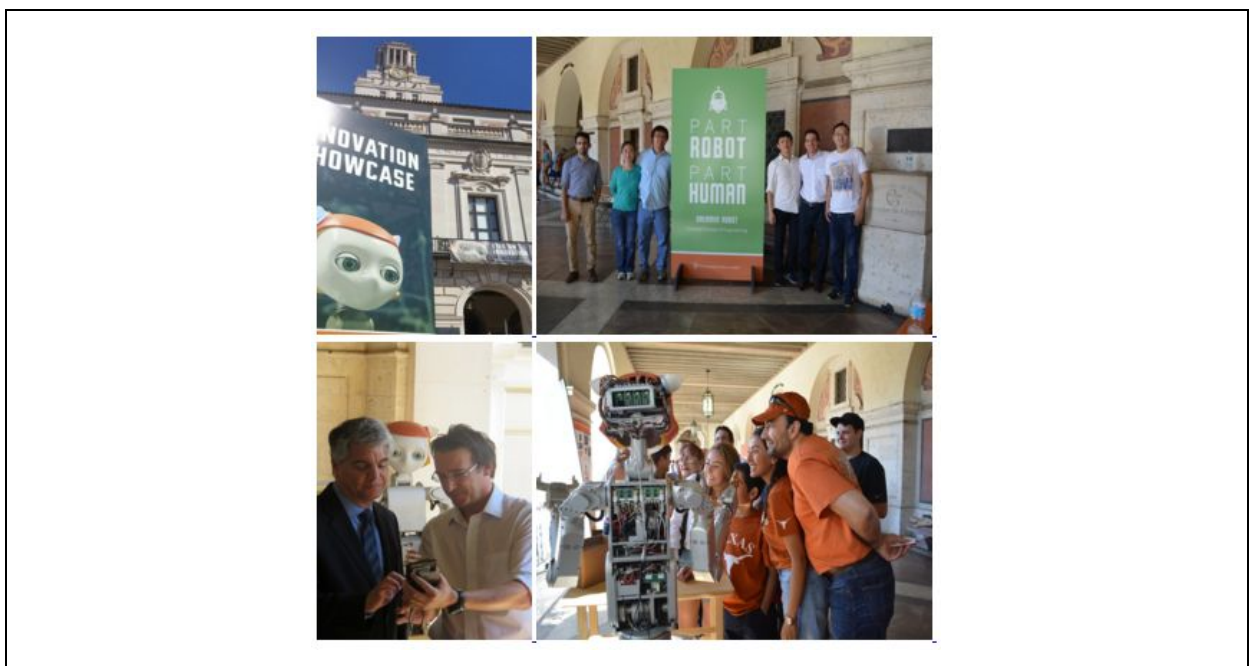

**Figure 5.** CARL being demonstrated to visitors at the 2015 Texas Tribune Festival, which was held on the University of Texas at Austin campus.

We also used CARL to repeatedly perform a dual-arm manipulation task. Figure 6 contains screenshots from a video showing a human operator using both of Dreamer's arms to pick up a bottle. Specifically, the right hand is used to push the bottle towards the left gripper, which then grabs the bottle. The operator is located about 1.6 miles away and cannot see the robot directly. A full video is available here: https://youtu.be/DMeVcJRKqnk. This demonstrates the feasibility of CARL's software architecture.

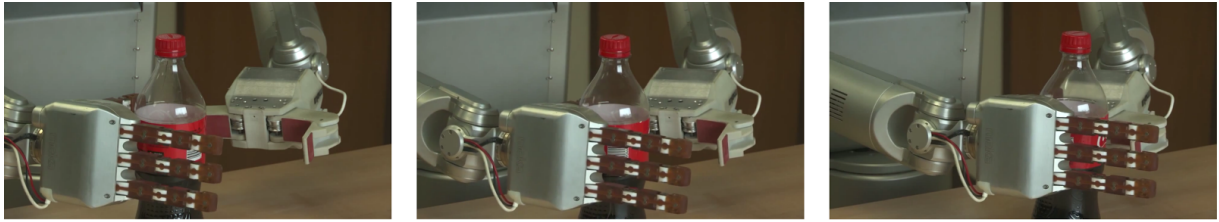

**Figure 6.** Snapshots from a video showing CARL being used to make Dreamer perform a manipulation task involving both end effectors. The right hand first pushes the bottle towards the left gripper, which then grabs the bottle. The operator is remotely controlling Dreamer using a smartphone 1.6 miles away.

## 7. Conclusions and Future Work

We presented and demonstrated a proof of concept software architecture for the Cloud-based Advanced Robotics Laboratory (CARL), which enables operators to access and control humanoid robots over the Internet. Its key distinguishing features include the use of a whole body controller to allow the user to work in a lower dimensional space, and a novel user interface that is accessible via any modern web browser including those on smartphones. We demonstrated the system's feasibility and ease of use by controlling Dreamer, a humanoid robot, from afar and allowing untrained members of the general public take immediate control of the robot and accomplish a dual arm manipulation task.

There are many areas of future work enabled by CARL. As previously mentioned, the current user interface is a proof of concept and can be enhanced by integrating more advanced 3D point cloud sensors [Yonekura 2012] and user input devices like 3D mouse, joysticks, or even accelerometers and gyroscopes of smartphones [Parga 2013] etc. But probably the most exciting future work is the integration of vast amounts of computational resources that are available in the cloud. Although the unpredictable latency and limited bandwidth over the Internet cause WBC almost impossible to process in the cloud, these computational resources can be leveraged for advanced artificial intelligence [Pratt 2015]. It could also be used to integrate humanoid robots with the Internet of Things (IoT) [Al-Fuqaha 2015] like Google's Brillo and Weave.

Cloud-based applications and services naturally facilitate collaboration. For example, multiple people can simultaneously edit the same Google Doc, or modify the content of a local folder that is shared and synchronized by Dropbox. In the future, we would like to extend CARL to support collaboration. This can take the form of multiple operators controlling multiple robots achieve a shared objective. It could also take the form of multiple operators controlling different aspects of a single robot. For example, it may be difficult for a single operator to focus both on navigating a rugged terrain and tracking a mobile target. This is analogous to how a team of operators on Earth controll just two rovers on Mars.

On the other hand, humanoid robots will likely one day outnumber humans which means each person may need to handle multiple humanoid robots. A cloud-based extension of CARL is to apply networked robotics [Sanfeliu 2008, Cardozo 2010] to construct a management system for humanoid robots. Networked robotics replace the human operator with a machine and increasing the scale of connectivity between devices. Specifically, instead of a single operator controlling a single robot, networked robotics consists of numerous robots collaborating both with each other and with IoT devices. Coordination can be done in a centralized or distributed manner. Examples of centralized networked robotic systems include Amazon Robotics' mobile robotic warehouse fulfillment systems [Wurman 2008, Guizzo 2008]. Distributed networked robotic systems are still being researched [Rubenstein 2012]. They are sometimes called swarm robotics since the overall and desired behavior of the system is emergent and the number of robots and devices involved resembles a swarm [Tan 2012, Rubenstein 2014]. The results of networked robotics may be applied on top of the current CARL system to, for example, coordinate a team of humanoid robots.

As previously mentioned, different types of whole body controllers exist and vary in their capabilities. In addition, each of these whole body controllers also have an infinite number of configurations. Different types of controllers and configurations are needed at different time since they offer different control laws and levels of performance. The controller performances can be mapped [Ashok 2013] based on the task contents and controller configurations. A future direction of CARL is to enable these controllers, their configurations, and performance maps to be stored in the cloud and dynamically deployed onto the humanoid robots based on the context. The user interface should visualize these performance maps during the teleoperation so that inexperienced operator can have a clear physical interpretation [Tisius 2009] of the whole body control laws. Constructing all the performance maps can be a challenge work but can be facilitated by CARL's user data stored in the cloud. Also, downloading the controllers and their configurations onto the robot is necessary to ensure real-time 1kHz servo frequencies. However, it would be interesting to investigate how parts of the whole body controller that are less latency-sensitive like updates to the robot model can be moved into the cloud.

Furthermore, we would like to enhance CARL to support in-browser software development and remote code deployment so that researchers will be able to test their codes with greater access of state-of-art humanoid robotic hardware. CARL's web portal should allow the researchers to write codes, compile it, and deploy it onto a selected humanoid robot. The researchers should be able to reprogram the motion planners, whole-body controllers, and even firmware within each joint, all from within CARL portal. Ultimately, we would like CARL to offer a Platform-as-a-Service (PaaS) to humanoid robotics developers. In order to achieve this goal, CARL should support all modern code development techniques like distributed version control, issue trackers, wikis, unit tests, continuous integration systems, and automatic documentation generators. Safety of operation will also be a huge concern for using this platform. Research can be conducted to identify models which define acceptable robot states. These models should be then embedded into CARL so that it can monitor the robot's current state to ensure it falls within the set of acceptable states. When the robot's state falls outside of the model, CARL should terminate user's experiment.

# References

1. E. Ackerman and E. Guizzo, "DARPA Robotics Challenge: Amazing Moments, Lessons Learned, and What's Next," IEEE Spectrum, June 11, 2015. Available online.
2. H. Ahn, H. Kim, Y. Oh, S. Oh (2014). Smartphone-Controlled Telerobotic Systems. In Cyber-Physical Systems, Networks, and Applications (CPSNA), 2014 IEEE International Conference on (pp. 77-80). IEEE.
3. A. Al-Fuqaha, M. Guizani, M. Mohammadi, M. Aledhari, and M. Ayyash, "Internet of Things: A Survey on Enabling Technologies, Protocols, and Applications," in Communications Surveys & Tutorials, IEEE , vol.17, no.4, pp.2347-2376, Fourthquarter 2015.
4. B. Annable, D. Budden, A. Mendes, "NUbugger: A Visual Real-Time Robot Debugging System," RoboCup 2013: Robot World Cup XVII, Vol. 8371, pp. 544-551.
5. R. Arumugam, V. R. Enti, L. Bingbing, W. Xiaojun, K. Baskaran, F. F. Kong, A. S. Kumar, K. D. Meng, and G. W. Kit, "DAvinCi: A Cloud Computing Framework for Service Robots," in IEEE Int. Conf. on Robotics and Automation (ICRA), 2010.
6. P. Ashok, and D. Tesar, (2013). The need for a performance map based decision process. Systems Journal, IEEE, 7(4), pp.616-631.
7. S. Bellens, K. Buys, N. Vanthienen, T. De Laet, R. Smits, M. Klotzbücher, W. Decré, H. Bruyninckx, J. De Schutter, "Haptic coupling with augmented feedback between the KUKA youBot and the PR2 robot arms," in Proc. of the International Conference on Intelligent Robots and Systems (IROS), San Francisco, California, 25-30 September, 2011.
8. R. L. Brown, H. L. Helton, A. C. Williams, M. T. Shrove, M. Milosevic, E. J. D. Coe, and J. Kulick, (2013). Android control application for Nao humanoid robot. In Proceedings of the International Conference on Frontiers in Education: Computer Science and Computer Engineering (FECS) (p. 1). The Steering Committee of The World Congress in Computer Science, Computer Engineering and Applied Computing (WorldComp).
9. M. Bryan, J. Green, M. Chung, L. Chang, R. Scherer, J. Smith, R. P. N. Rao (2011), "An Adaptive Brain-Computer Interface for Humanoid Robot Control," in Proc. of IEEE/RAS Int. Conf. on Humanoid Robots, 2011, pp. 199-204.
10. K. Buys, S. Bellens, D. Vanthienen, W. Decré, M. Klotzbücher, T. De Laet, R. Smits, H. Bruyninckx, J. De Schutter, "Haptic coupling with the PR2 as a demo of the OROCOS - ROS - Blender integration," IROS PR2 Workshop. San Francisco, California, 25-30 September 2011.
11. E. Cardozo, E. Guimaraes, L. Rocha, R. Souza, F. Paolieri, and F. Pinho, "A Platform for Networked Robotics," in IEEE / RSJ Int. Conf. on Intelligent Robots and Systems, Oct. 18-22, 2010, Taipei, Taiwan.
12. J. S. Castellanos A., L. E. Solaque G., M. M. Arteche (2008), "Model and Implementation of Master-Slave Basic Goniometric System for Real Time Control of a Mini Humanoid

Robot," in Proc. of Electronics, Robotics, and Automotive Mechanics Conference, 2008, pp. 598-603.
13. M. J.-Y. Chung, M. Forbes, M. Cakmak, and R. P. Rao, "Accelerating Imitation Learning Through Crowdsourcing," Journal of Human-Robot Interaction, Vol. 3, No. 2, pp. 25-49, 2014.
14. C. Crick, S. Osentoski, G. Jay, and O. C. Jenkins, "Human and Robot Perception in Large-Scale Learning from Demonstration," In Proc. of the 6th ACM / IEEE International Conference on Human-Robot Interaction, 2011.
15. J. J. Dudley, "Enhancing Awareness to Support Teleoperation of a Bulldozer," Master's Thesis, The Univ. of Queensland, 2014.
16. A. Finke, B. Rudgawawis, H. Jakusch, H. Ritter (2012), "Towards multi-user brain-robot interfaces for humanoid robot control," in Proc. of IEEE-RAS Int. Conf. on Humanoid Robots, 2012, pp. 532-537.
17. A. Finke, N. Hachmeister, H. Riechmann, and H. Ritter (2013), "Thought-Controlled Robots - Systems, Studies, and Future Challenges," in Proc. of IEEE Int. Conf. on Robotics and Automation (ICRA), 2013, pp. 3403-3408.
18. C.-L. Fok, G. Johnson, J.D. Yamokoski, A. Mok, and L. Sentis, "ControlIt! - A Middleware for Whole Body Operational Space Control," in Int. Journal of Humanoid Robotics, 2015.
19. K. Goldberg and R. Siegwart, Ed., "Beyond Webcams An Introduction to Online Robots," MIT Press, Cambridge, MA, November 2001.
20. K. Goldberg, Ed., "The Robot in the Garden: Telerobotics and Telepistemology in the Agent of the Internet," MIT Press, Cambridge, MA, November 2001.
21. E. Guizzo, "Three Engineers, Hundreds of Robots, One Warehouse," IEEE Spectrum, July 2008, pp. 26-34.
22. S. Hart, P. Dinh, J. D. Yamokoski, B. Wightman, and N. Radford (2014), "Robot Task Commander: A Framework and IDE for Robot Application Development," in Proc. of IEEE/RSJ Int. Conf. on Intelligent Robots and Systems (IROS), 2014, pp. 1547-1554.
23. S. Hart, P. Dinh, K. A. Hambuchen (2014), "Affordance Templates for Shared Robot Control," in Proc. of AAAI Symposium, 2014, pp. 81-82.
24. S. Hart, P. Dinh, K. A. Hambuchen (2015), "The Affordance Template ROS Package for Robot Task Programming," in Proc. of IEEE Int. Conf. on Robotics and Automation (ICRA), 2015, pp. 6227 - 6234.
25. G. Hu, W. P. Tay, and Y. Wen, "Cloud robotics: architecture, challenges and applications," in *Network, IEEE* , vol.26, no.3, pp.21-28, May-June 2012
26. B. Kehoe, A. Matsukawa, S. Candido, J. Kuffner, and K. Goldberg, "Cloud-based Robot Grasping with Google Object Recognition Engine," in Proc. of the IEEE Int. Conf. on Robotics and Automation (ICRA), 2013.
27. B. Kehoe, S. Patil, P. Abbeel, and K. Goldberg, "A Survey of Research on Cloud Robotics and Automation," in *Automation Science and Engineering, IEEE Transactions on* , vol.12, no.2, pp.398-409, April 2015
28. M. Klotzbucher and H. Bruyninckx, "Coordinating Robotic Tasks and Systems with rFSM Statecharts," in Journal of Software Engineering for Robotics, 1(3), January 2012, pp. 28-56.

29. A. Koubaa, "ROS As a Service: Web Services for Robot Operating System," Journal of Software Engineering for Robotics, 6(1), December, 2015.
30. S. Kuindersma, F. Permenter, and R. Tendrake (2014), "An Efficiently Solvable Quadratic Program for Stabilizing Dynamic Locomotion," ICRA 2014.
31. S. Kuindersma, R. Deits, M. Fallon, A. Valenzuela, H. Dai, F. Permenter, T. Koolen, P. Marion, R. Tedrake (2015), "Optimization-based Locomotion Planning, Estimation, and Control Design for the Atlas Humanoid Robot," in Autonomous Robots, July 2015.
32. E. S. Neo, K. Yokoi, S. Kajita, and K. Tanie (2007), "Whole-Body Motion Generation Integrating Operator's Intention and Robot's Autonomy in Controlling Humanoid Robots," in IEEE Trans. on Robotics, Vol. 23, N. 4, August 2007, pp. 763-775.
33. T. Nishiyama, H. Hoshino, K. Sawada, Y. Tokunaga, H. Shinomiya, M. Yoneda, I. Takeuchi, Y. Ichige, S. Hattori, and A. Takanishi (2003), "Development of User Interface for Humanoid Service Robot System," in Proc. of IEEE Int. Conf. on Robotics & Automation, 2003, pp. 2979 - 2984.
34. Y. Ogawa, G. Venture, C. Ott (2014), "Dynamic Parameters Identification of a Humanoid Robot Using Joint Torque Sensors and/or Contact Forces," in IEEE-RAS Int. Conf. on Humanoid Robots, 2014, pp. 457-462.
35. S. Osentoski, B. Pitzer, C. Crick, G. Jay, S. Dong, D. H. Grollman, H. B. Suay, and O. C. Jenkins, "Remote robotic laboratories for learning from demonstration -- Enabling User Interaction and Shared Experimentation," International Journal of Social Robotics," vol. 4, no. 4, pp. 449-461, 2012.
36. C. Parga, X. Li, W. Yu, (2013). Tele-manipulation of robot arm with Smartphone. In Resilient Control Systems (ISRCS), 2013 6th International Symposium on (pp. 60-65). IEEE.
37. J. Peters, S. Vijayakumar, S. Schaal (2003), "Reinforcement Learning for Humanoid Robots," in IEEE-RAS Int. Conf. on Humanoid Robots.
38. O. Pettersson (2014), "Execution Monitoring in Robotics: A Survey," in Robotics and Autonomous Systems 53, 20015, pp. 73-88.
39. B. Pitzer, S. Osentoski, G. Jay, C. Crick, and O. Jenkins, "PR2 Remote Lab: An Environment for Remote Development and Experimentation," in IEEE International Conference on Robotics and Automation (ICRA), May 2012.
40. V. Poorten, E. Demeester, P. Lammertse, "Haptic Feedback for Medical Applications, A Survey," Proceedings Actuator 2012.
41. G. Pratt, "Is a Cambrian Explosion Coming for Robotics?," in Journal of Economic Perspectives, 29(3): 51-60, 2015.
42. M. Quigley, B. Gerkey, K. Conley, J. Faust, T. Foote, J. Leibs, E. Berger, R. Wheeler, A. Ng (2009), "ROS: An Open-source Robot Operating System," in ICRA Workshop on Open Source Software.
43. L. Riazuelo, M. Tenorth, D. Di Marco, M. Salas, D. Galvez-Lopez, L. Mosenlechner, L. Kunze, M. Beetz, J. D. Tardos, L. Montano, J. M. Martinez Montiel, "RoboEarth Semantic Mapping: A Cloud Enabled Knowledge-Based Approach," Automation Science and Engineering, IEEE Transactions on, On page(s): 432 - 443 Volume: 12, Issue: 2, April 2015.

44. B. Ross, J. Bares, D. Stager, L. Jackel, M. Perschbacher, "An Advanced Teleoperation Testbed," in Proc. of the 6th International Conference on Field and Service Robotics - FSR 2007, Jul. 2007, Chamonix, France. Springer, 42, 2007, Springer Tracts in Advanced Robotics; Field and Service Robotics.
45. M. Rubenstein, C. Ahler, R. Nagpal, "Kilobot: A Low Cost Scalable Robot System for Collective Behaviors," in ICRA 2012.
46. M. Rubenstein, A. Cornejo, R. Nagpal, "Programmable Self-Assembly in a Thousand-Robot Swarm," Science, 345(6198), pp. 795-799, Aug. 15, 2014.
47. A. Sanfeliu, N. Hagita, A. Saffiotti, "Network Robot Systems," in Journal of Robotics and Autonomous Systems, 56(10), October 2008, pp. 793-797.
48. L. Sentis, (2007), "Synthesis and Control of Whole-Body Behaviors in Humanoid Systems," Stanford University, PhD Thesis, supervised by Oussama Khatib.
49. L. Sentis, "Compliant Control of Whole-Body Multi-Contact Behaviors in Humanoid Robots", Motion Planning for Humanoid Robots, Springer Global Editorial, Chapter 2, August 2010, pp. 29-63
50. L. Sentis, J. Petersen, R. Philippsen, "Implementation and stability analysis of prioritized whole-body compliant controllers on a wheeled humanoid robot in uneven terrains," International Journal of Autonomous Robots, 35(4), pp. 301-319, 2013.
51. N. E. Sian, K. Yokoi, S. Kajita, and K. Tanie (2004), "A Framework for Remote Execution of Whole Body Motions for Humanoid Robots," in Proc. of Humanoids 2004, pp. 608-626.
52. A. Stumpf, S. Kohlbrecher, D. C. Conner, and O. v. Stryk (2014), "Supervised Footstep Planning for Humanoid Robots in Rough Terrain Tasks using a Black Box Walking Controller," in Proc. of IEEE-RAS Int. Conf. on Humanoid Robots, 2014, pp. 287-294.
53. T. Takubo, K. Nishii, K. Inoue, Y. Mae, T. Arai (2004), "Marionette System for Operating and Displaying Robot Whole-Body Motion-Development of Similar Humanoid-Type Device," in Proc. of IEEE/RSJ Int. Conf. on Intelligent Robots and Systems, 2004, pp. 509-514.
54. T. Takubo, K. Inoue, T. Arai, and K. Nichii (2006), "Wholebody Teleoperation for Humanoid Robot By Marionette System," in Proc. of IEEE/RSJ Int. Conf. on Intelligent Robots and Systems, 2006, pp. 4459-4465.
55. Y. Tan, Y. Shi, and Z. Ji, "Advances in Swarm Intelligence," in Third Int. Conf. on Swarm Intelligence, June 17-20, 2012, proceedings, part 1.
56. K. Taylor and J. Trevelyan, "A Telerobot On the world Wide Web," in Proceedings of the National Conference of the Australian Robot Association, Melbourne, Australia, July 1995.
57. M. Tisius, M. Pryor, C. Kapoor, and D. Tesar, (2009). An empirical approach to performance criteria for manipulation. Journal of Mechanisms and Robotics, 1(3), p.031002.
58. R. Toris, D. Kent, and S. Chernova, "The Robot Management System: A Framework for Conducting Human-Robot Interaction Studies Through Crowdsourcing," Journal of Human-Robot Interaction, Vol. 3, No. 2, pp. 25-49, 2014.

59. R. Toris, J. Kammerl, D. Lu, J. Lee, O. C. Jenkins, S. Osentoski, M. Wills, and Sonia Chernova, "Robot Web Tools: Efficient Messaging for Cloud Robotics," in Proceedings of the IEEE/RSJ International Conference on Intelligent Robots and Systems (IROS), 2015
60. D. Vanthienen, T. De Laet, W. Decré, R. Smits, M. Klotzbücher, K. Buys, S. Bellens, L. Gherardi, H. Bruyninckx, J. De Schutter (2011), "iTaSC as a unified framework for task specification, control, and coordination, demonstrated on the PR2," in proc. of the IEEE/RSJ International Conference on Intelligent Robots and Systems, San Francisco, 25-30 September 2011.
61. D. Vanthienen, M. Klotzbücher, J. De Schutter, T. De Laet, and H. Bruyninckx (2013), "Rapid Application Development of Constraint-based task Modelling and Execution Using Domain Specific Languages," in Proceedings of the 2013 IEEE/RSJ International Conference on Intelligent Robots and Systems, Tokyo, 3-8 November 2013 (pp. 1860-1866).
62. M. Waibel, M. Beetz, J. Civera, R. D'Andrea, J. Elfring, D. Galvez-Lopez, K. Haussermann, R. Janssen, J. M. M. Montiel, A. Perzylo, B. Schieble,m M. Tenorth, O. Zweigle, and R. van de Molengraft, "RoboEarth - A World Wide Web for Robots," IEEE Robotics and Automation Magazine, June 2011, pp. 69-82.
63. Website - RoboEarth, http://roboearth.org/, last accessed December 21, 2015.
64. Website - DRC, DARPA Robotics Challenge, http://theroboticschallenge.org/, last accessed January 1, 2016.
65. Website - AWS, Amazon Web Services, last accessed January 2, 2016.
66. Website - IEEE RAS Whole-body Control Technical Committee, http://www.ieee-ras.org/whole-body-control, last accessed January 3, 2016.
67. Website - Mars Exploration Rovers,
68. B. Willaert, B. Corteville, H. Bruyninckx, H. V. Brussel, and E. B. V. Poorten, "Mechatronic Design Optimization of a Teleoperation System based on Bounded Environment Passivity," in EuroHaptics, 2010.
69. P. Wurman, R. D'Andrea, and M. Mountz, "Coordinating Hundreds of Cooperative, Autonomous Vehicles in Warehouses," AI Magazine, Vol. 29, No. 1, Spring 2008.
70. K. Yonekura, S. Nakaoka, K. Yokoi (2012), "Whole-body Motion Input Method for Bipedal Humanoid Robot with Support Leg Detection," in Proc. of IEEE-RAS Int. Conf. on Humanoid Robots, 2012, pp. 853-858.